\documentclass[letterpaper]{article} 
\usepackage{aaai24}  
\usepackage{times}  
\usepackage{helvet}  
\usepackage{courier}  
\usepackage[hyphens]{url}  
\usepackage{graphicx} 
\usepackage{natbib}  
\usepackage{caption} 
\usepackage{algorithm}
\usepackage{algorithmic}

\usepackage{newfloat}
\usepackage{listings}
\DeclareCaptionStyle{ruled}{labelfont=normalfont,labelsep=colon,strut=off} 
\floatstyle{ruled}
\newfloat{listing}{tb}{lst}{}
\floatname{listing}{Listing}
\usepackage{booktabs}
\usepackage{multirow}
\usepackage{adjustbox}
\usepackage{microtype}
\usepackage{amsmath}
\usepackage{amssymb}

\title{Fine-Grained Prototypes Distillation for Few-Shot Object Detection}
\author{
    Zichen Wang, Bo Yang\thanks{Corresponding author.}, Haonan Yue, Zhenghao Ma
}
\affiliations{
    School of Automation, Northwestern Polytechnical University, Xi'an, China\\

    \{wangchen1801, hnyue, mazh0819\}@mail.nwpu.edu.cn, 
    byang@nwpu.edu.cn
}

\usepackage{bibentry}

\begin{document}

\maketitle

\begin{abstract}
	Few-shot object detection (FSOD) aims at extending a generic detector for novel object detection with only a few training examples. It attracts great concerns recently due to the practical meanings. 
	Meta-learning has been demonstrated to be an effective paradigm for this task. In general, methods based on meta-learning employ an additional support branch to encode novel examples (a.k.a. support images) into class prototypes, which are then fused with query branch to facilitate the model prediction. However, the class-level prototypes are difficult to precisely generate, and they also lack detailed information, leading to instability in performance.
	New methods are required to capture the distinctive local context for more robust novel object detection. To this end, we propose to distill the most representative support features into fine-grained prototypes. These prototypes are then assigned into query feature maps based on the matching results, modeling the detailed feature relations between two branches. This process is realized by our Fine-Grained Feature Aggregation (FFA) module. 
	Moreover, in terms of high-level feature fusion, we propose Balanced Class-Agnostic Sampling (B-CAS) strategy and Non-Linear Fusion (NLF) module from differenct perspectives. They are complementary to each other and depict the high-level feature relations more effectively. Extensive experiments on PASCAL VOC and MS COCO benchmarks show that our method sets a new state-of-the-art performance in most settings.
	Our code is available at https://github.com/wangchen1801/FPD. 
\end{abstract}

\section{Introduction}
Object detection is a fundamental task in computer vision and the methods based on deep learning have been well established over the past few years~\cite{redmon2016you,ren2017faster,carion2020end,liu2016ssd}.
While remarkable achievements have been made, most of them require a large amount of labeled data to obtain a satisfactory performance, otherwise they are prone to overfitting and hardly generalize to the unknow data. 

\begin{figure}
	\centering
	\includegraphics[width=0.91\linewidth]{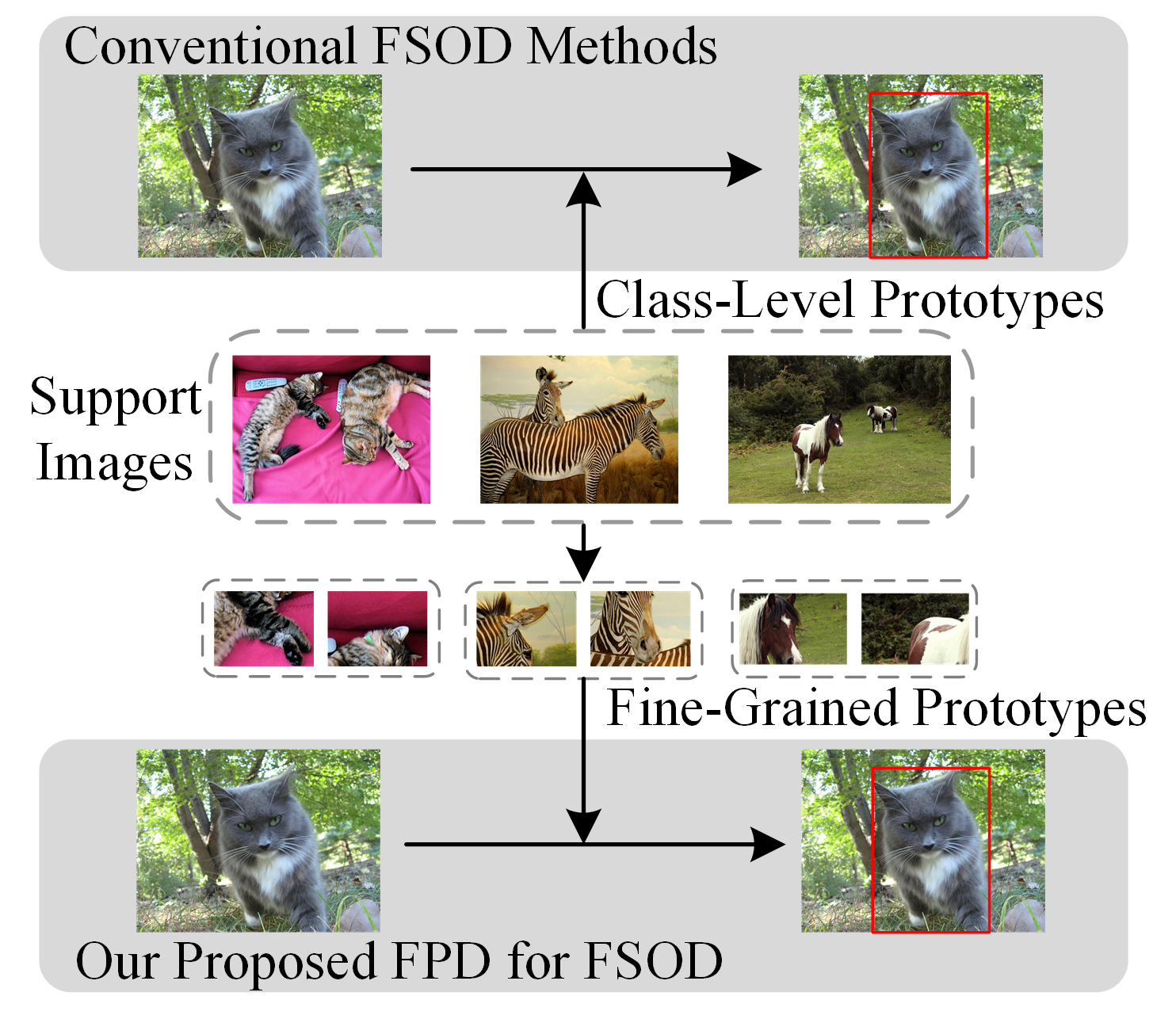}
	\caption{Overview of the proposed method, which we denote as FPD. In addition to class-level prototypes, we distill representative detailed features into fine-grained prototypes, enabling more robust novel object detection.}
	\label{fig:overview}
\end{figure}

\begin{figure*}[t]
	\centering
	\includegraphics[width=\textwidth]{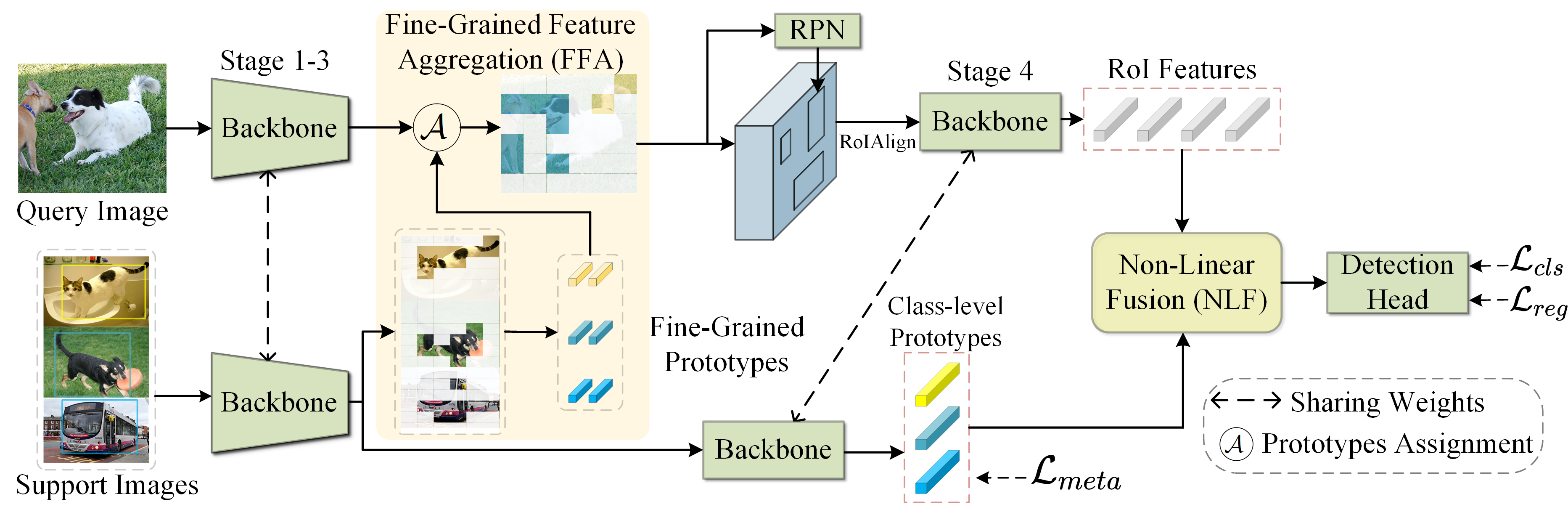}
	\caption{The overall architecture of our method. FFA and NLF are proposed to improve the performance. }
	\label{fig:architecture}
\end{figure*}

Few-shot object detection (FSOD) is a more challenging task to detect object specially in data-scarce scenarios. FSOD assumes that there are sufficient amount of examples for base classes while only k-shot examples for each novel class. Therefore, the key question is how to transfer the knowledge learnt from base classes to the novel classes. 
\textbf{Transfer learning} based methods~\cite{wang2020frustratingly,cao2021few,qiao2021defrcn} focus on fine-tuning the model more effectively. They use the same architecture as generic object detection, additionally with advanced techniques such as parameter freezing and gradient decoupling to improve performance. 
\textbf{Meta-learning} based methods~\cite{kang2019few,wang2019meta,yan2019meta,han2023vfa}, 
instead, follow the idea: learn how to learn the new tasks rapidly. As illustrated in Figure~\ref{fig:architecture}, an additional support branch is incorporated to encode support images into class-level prototypes, which function as dynamic parameters to interact with the query branch. In this way, the connections between novel examples and the model predictions are enhanced, thereby improving the generalization ability and learning the new tasks more quickly. 

This work studies the meta-learning based FSOD and aims at realizing a more effective method. In general, features from the two branches are fused on top of the framework to make the final prediction~\cite{kang2019few,yan2019meta,xiao2020few}, 
while most of the layers are separated and do not exchange information. This hinders the model from learning the correlations among detailed features especially in data-scarce scenarios. 

DCNet~\cite{hu2021dense} proposes to directly match the mid-level support features into query features in a pixel-wise manner, which enables the relation modeling of detailed local context. However, this approach has its limitations in terms of effect and implementation. \textbf{First}, the mid-level features with an extensive range of patterns are intricate and complex, thus the model might struggle to capture the most critical details. 
\textbf{Second}, directly matching between dense feature maps is inefficiency and will cost more computational resources.
\textbf{Third}, this approach has difficulty in transitioning seamlessly from the training phase to the testing phase, as it can not integrate the mid-level support features across different shots to boost the performance.

To address the aforementioned issues, we propose a novel Fine-Grained Feature Aggregation (FFA) module to aggregate the mid-level features. As illustrated in Figure~\ref{fig:FFA}, different from DCNet, we propose to distill features into fine-grained prototypes. These prototypes, which reside in a highly refined and reduced feature space, embody the most distinctive and representative details of the support images. Specifically, we employ a set of embeddings following the object queries in DETR~\cite{carion2020end} to distill prototypes. Rather than being encoded with positional information and representing specific objects, the embeddings here function within the feature space and thereby are denoted as feature queries. We give each class a unique set of feature queries to distill prototypes independently. It can avoid confusion and is a key factor for our method to work. 
The distilled prototypes are then assigned into query feature map based on the matching results, modeling the fine-grained relations and highlighting the features with similar details. 

The proposed FFA enables a more effective feature aggregation by focusing on the key information encapsulated within prototypes. This method also reduces the computational complexity by avoiding the directly matching between dense feature maps. Furthermore, it can naturally transition into the testing phase through a weighted sum of prototypes across different shots, preserving the full potential derived from the training phase. 

In terms of high-level feature aggregation, we revisit the previous methods and propose two improvements from different perspectives.
\textbf{First}, we propose Balanced Class-Agnostic Sampling (B-CAS) strategy to control the ratio of support classes aggregated with query features. 
Meta R-CNN~\cite{yan2019meta} adopts a simple class-specific aggregation scheme where 
only the features having the same classes are aggregated. 
While VFA~\cite{han2023vfa} proposes a class-agnostic aggregation scheme which randomly selects the support classes to reduce class bias. Our insight is that different support classes are served as positive and negative samples, thereby the balanced sampling is required to keep the most important positive samples from being overwhelmed.
\textbf{Second}, many works~\cite{kang2019few,yan2019meta,han2023vfa} employ element-wise multiplication to explore the relations within the same classes. However, it is not compatible with our proposed B-CAS which incorporates the feature aggregation between different classes. To solve this issue, we propose a stronger Non-Linear Fusion (NLF) module motivated by~\cite{han2022meta,xiao2020few} to fuse features more effectively. Our contributions can be summarized as follows: 

\begin{itemize}
	\item We propose to distill support features into fine-grained prototypes before being integrated into query feature maps, which can help the model grasp the key information. They are implemented in the Fine-Grained Feature Aggregation (FFA) module. 
	\item We propose Balanced Class-Agnostic Sampling (B-CAS) strategy and Non-Linear Fusion (NLF) module. They are complementary to each other and can fuse high-level features more effectively.
	\item Extensive experiments illustrate that our method 
	significantly improves the performance and achieves state-of-the-art results on the two widely used FSOD benchmarks. 
\end{itemize}

\section{Related Works}
\subsection{General Object Detection}
Deep learning based object detection has been extensively studied in recent years. The well-established object detectors can be categorized into one-stage and two-stage methods. One-stage detectors~\cite{redmon2016you,liu2016ssd,lin2017focal} directly make predictions upon the CNN feature maps. While two-stage detectors~\cite{ren2017faster,he2017mask} additionally employ a Region Proposal Network (RPN) to generate object proposals, which will be further refined into the final predictions. Both of them require the predefiend dense anchors to generate candidates. 

Recently, anchor-free detectors DETR~\cite{carion2020end} and Deformable DETR~\cite{zhu2020deformable} have been developed and are drawing more attention. They use a CNN backbone combining with Transformer encoder-decoders~\cite{vaswani2017attention} for end-to-end object detection. 
A set of object queries are proposed to replace the anchor boxes. They will be refined into the detected objects layer by layer through Transformer decoders. 

We employ the two-stage Faster R-CNN~\cite{ren2017faster} framework to build our FSOD detector, and draw inspirations from DETR~\cite{carion2020end} into our approach.

\subsection{Few-Shot Object Detection}
Few-Shot Object Detection (FSOD), which studies the detection task in data-scarce situations, has been attracting an increased interest recently. 
LSTD~\cite{chen2018lstd} first proposes a transfer learning based approach to detect novel objects in a FSOD data setting. TFA~\cite{wang2020frustratingly} utilizes a cosine similarity based classifier and only fine-tunes the last layer with novel examples, achieving a comparable results with other complex methods. DeFRCN~\cite{qiao2021defrcn} employs advanced gradient decoupling technique into the Faster R-CNN framework and intergrates an offline prototypical calibration block to refine the classification results, which achieves an impressive performance. 

The meta-learning is also a promising paradigm for FSOD. FSRW~\cite{kang2019few} proposes to re-weight the YOLOv2 feature maps along channel dimension using proposed reweighting vectors, which can highlight the relevant features. Meta R-CNN~\cite{yan2019meta} adopts the Faster R-CNN framework to build a two-branch based siamese network. It processes query and support images in parallel to produce the Region of Interest (RoI) features and class prototypes, which are then fused to make predictions. Instead of learning a softmax-based classifier for all classes, \cite{han2022meta} constructs a meta-classifier through feature alignment and non-linear matching. It calculates the similarity between query-support feature maps, producing binary classification results for novel classes. VFA~\cite{han2023vfa} introduces variational feature learning into Meta R-CNN, further boosting its performance. Recently, there are some works incorporate meta-learning into other advanced frameworks. Meta-DETR~\cite{zhang2022metadetr} employs Deformable DETR~\cite{zhu2020deformable} to build a few-shot detector. \cite{han2022fct} utilizes PVT~\cite{wang2021pyramid} to construct a fully cross transformer for few-shot detection. They all achieve remarkable results. 

A two stage training paradigm has been widely adopted in both transfer learning and meta-learning based methods due to its effectiveness. At the base training stage, the model is trained on abundant base class examples. While at the fine-tuning stage, the model is fine-tuned only with $K$-shot examples for each base and novel class.

Our approach is based on Meta R-CNN and we propose to distill fine-grained prototypes for effectively exploiting the relations between detailed features.

\section{Our Approach}
In this section, we first introduce the task definition and the overall architecture of our model. Then we will elaborate the fine-grained and high-level feature aggregation.

\subsection{Task Definition}
We adopt the standard FSOD setting following~\cite{kang2019few, wang2020frustratingly}. Specifically, given a dataset $\mathcal{D}$ with two sets of classes $C_{base}$ and $C_{novel}$, where each class in $C_{base}$ has abundant training data while each class in $C_{novel}$ has only $K$-shot annotated objects, FSOD aims at detecting the objects of $C_{base} \cup C_{novel}$ using the detector trained on $\mathcal{D}$. Please note that $C_{base} \cap C_{novel} = \varnothing$.

\subsection{The Model Architecture}
As illustrated in Figure~\ref{fig:architecture}, our model is based on Meta R-CNN, which is a siamese network with query branch and support branch that share a same backbone. Typically, we use the first three stages of ResNet-50/101 backbone~\cite{he2016resnet} to extract mid-level features for both query images and support images. Then our proposed FFA module is employed to distill the fine-grained prototypes and assign them into the query branch. Subsequently, we use the last stage (i.e. stage four) of the backbone to extract high-level features for both branches, which produces RoI features and class-level prototypes, respectively. They are further processed by the proposed NLF module, following by the detection head to make the final prediction. We would like to mention that the RPN is fed with the query features which have already interacted with the support branch. It gives the RPN more ability learning to identify the new instances.

\subsection{Fine-Grained Feature Aggregation}
The Fine-Grained Feature Aggregation (FFA) module is the key component of our proposed method, which is a class-agnostic aggregator that matchs all classes of support features into query features. It models inter-class relations in the early stage of the detection framework where the features are low-level and have more detailed information. Instead of directly performing feature matching, we propose to distill the representative support features into fine-grained prototypes. These prototypes are then assigned into query feature maps based on the matching results. 
FFA can help the model distinguish foreground from background and learn the similarities and differences between object classes. We will elaborate the prototypes distillation and feature assignment in the following subsections. We also discuss our strategy to transfer this method to novel classes, as well as test-time natural integration of prototypes across different shots.

\subsubsection{Prototypes Distillation}
Inspired by DETR, we incorporate a new component which is a set of learnable embeddings to distill prototypes. Different from object queries in DETR, which are encoded with positional information and are refined into a specific instance layer by layer, the embeddings here work as a guidance to refine the entire support feature space into a set of representative features. It can filter out the noise and ease the training. We refer to these embeddings as feature queries.

We employ the cross-attention mechanism to perform the prototypes distillation. Specifically, given a support feature map ${X_s} \in \mathbb{R}^{hw \times d}$ and a set of feature queries ${q} \in \mathbb{R}^{n \times d'}$, where $hw$ denote the height and width, $d$ and $d'$ is the feature dimension, and $n$ is the number of feature queries, the affinity matrix is calculated through a matching operation:
\begin{equation}
	A = softmax(\frac{q(X_{s}W)^{T}}{\sqrt{d'}})
\end{equation}
where $W$ is a linear projection to project $X_s$ in to the latent space with dimensionality $d'$, 
and the $softmax$ function is performed along $hw$ dimension. Subsequently, the fine-grained prototypes can be distilled from ${X_s}$ via:
\begin{equation}
	{p} = {A}{X_s} + {E}_{cls}
\end{equation}
where the affinity matrix is applied directly on the support feature map. We do not project $X_s$ to keep feature space the same. An additional class embedding ${E}_{cls}$ is added to retain the class information.

We would like to mention that each class has its exclusive feature queries. 
This is different from object queries in DETR and is a crucial factor for our method to work. It means that ${q}$ is the feature queries of one class and is part of ${Q} \in \mathbb{R}^{nc \times d}$, where ${Q}$ denotes the feature queries of all classes. This setting makes feature queries class-relevant and avoids them getting overwhelmed and confused by too many object classes.

\subsubsection{Prototypes Assignment}
We densely match the fine-grained prototypes into query feature map to achive the prototypes assignment. Considering that the background area should not be matched to any prototypes that represent salient object features, 
we incorporate a set of embeddings to serve as background prototypes.
We also use the cross-attention mechanism to assign prototypes. Specifically, given a query feature map ${X_q} \in \mathbb{R}^{HW \times d}$, prototypes assignment is performed via:
\begin{equation}
	{A'} = softmax(\frac{(X_qW')(PW')^T}{\sqrt{d'}})
\end{equation}
\begin{equation}
	{P} = concat({p}_{1}, {p}_{2}, ..., {p}_{c}, {p}_{bg})
\end{equation}
\begin{equation}
	{X_{q}'} = {X_q} + \alpha \cdot {A'}{P}
\end{equation}
where 
${P} \in \mathbb{R}^{(nc+n_{bg}) \times d}$ 
is the prototypes of $c$ support classes with additional $n_{bg}$ background classes, and ${W'}$ is a linear projection shared by $X_q$ and $P$ which projects them into the same latent space. The prototypes are assigned into query feature map based on the affinity matrix ${A'}$, which produces the aggregated query features. The $\alpha$ is a learnable parameter initialized as zero to help stabelize the training.

\begin{figure}[t]
	\centering
	\includegraphics[width=0.97\linewidth]{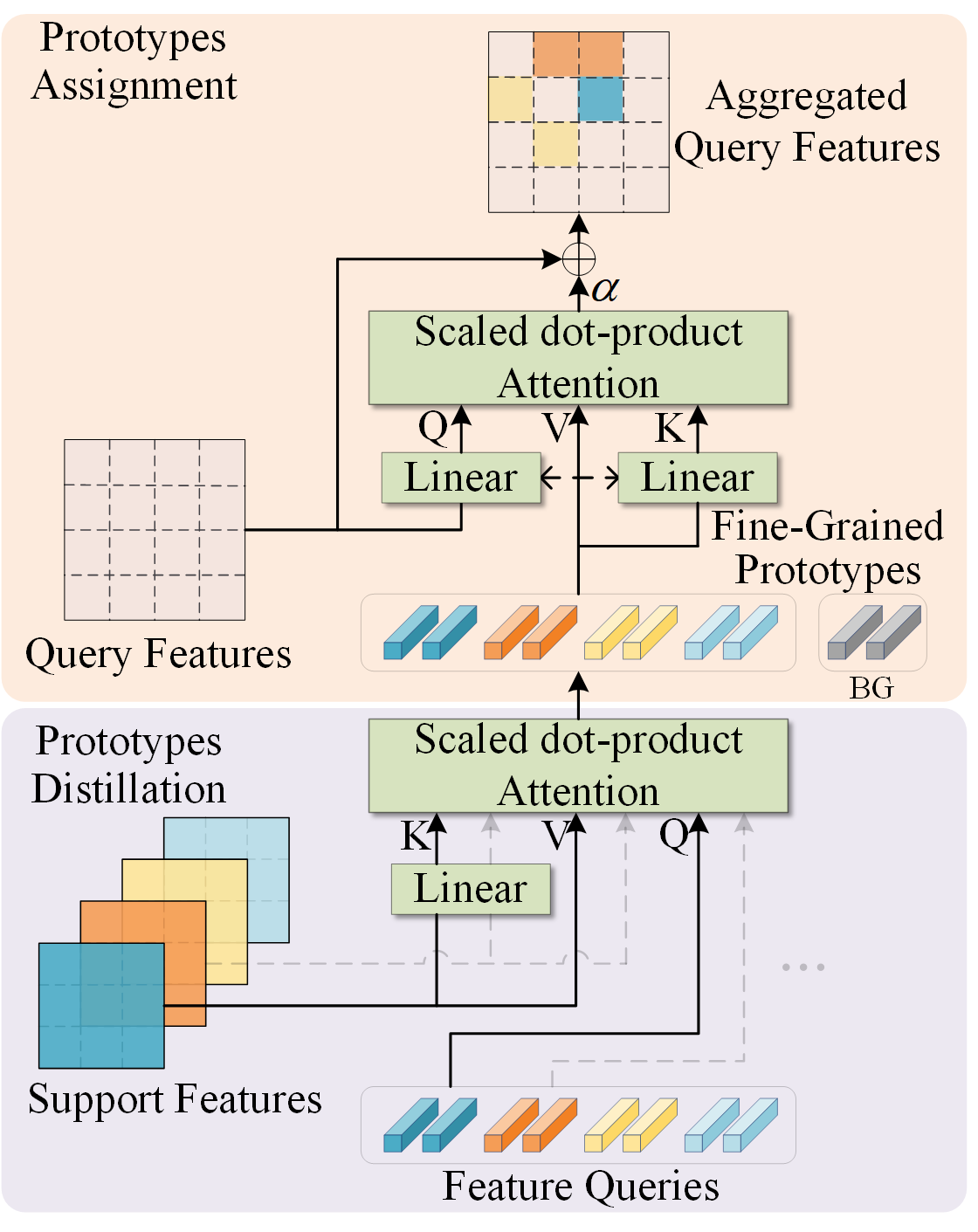}
	\caption{The architecture of the Fine-Grained Feature Aggregation (FFA) module. It can be divided into Prototypes Distillation and Prototypes Assignment. }
	\label{fig:FFA}
\end{figure}

\subsubsection{Transferring to Novel Classes}
At the base training stage, the feature queries of base classes are randomly initialized and well trained. However, at the fine-tuning stage, training the feature queries from scratch becomes challenging due to the limited novel class examples, which means that an effective knowledge transfer method is required. To address this issue, we propose to duplicate the most compatible feature queries from the base classes to serve as those in the novel classes. To be specific, given feature queries of base classes $Q \in \mathbb{R}^{nc \times d'}$ and support feature map of a novel class $X_{ns} \in \mathbb{R}^{hw \times d}$, the compatibility matrix and the $weight$ of each feature query can be obtained via:
\begin{equation}
	C = topk\left(Q(X_{ns}W)^T\right)
\end{equation}
\begin{equation}
	weight_i = \sum_{j=0}^{k} C_{ij}, i = 1, 2, ..., nc
\end{equation}
where $topk$ is performed along $hw$ dimension to filter out irrelevant locations. We select $n$ feature queries for each novel class based on the largest $weight$. Instead of sharing the same feture queries with base classes, they are created as a duplicate and can be trained independently.

\begin{table*}[ht]
	\centering
	\setlength{\tabcolsep}{4.6pt}
	\small
	\begin{tabular}{@{}l|ccccc|ccccc|ccccc@{}}
		\toprule
		\multirow{2}{*}{Method / shot} & \multicolumn{5}{c|}{Novel Set 1} & \multicolumn{5}{c|}{Novel Set 2} & \multicolumn{5}{c}{Novel Set 3} \\
		& 1 & 2 & 3 & 5 & 10 & 1 & 2 & 3 & 5 & 10 & 1 & 2 & 3 & 5 & 10 \\
		\midrule
		\multicolumn{16}{l}{\textit{Single run results:}} \\
		\midrule
		FSRW~\cite{kang2019few}&14.8&15.5&26.7&33.9&47.2&15.7&15.3&22.7&30.1&40.5&21.3&25.6&28.4&42.8&45.9 \\
		Meta R-CNN~\cite{yan2019meta}&19.9&25.5&35.0&45.7&51.5&10.4&19.4&29.6&34.8&45.4&14.3&18.2&27.5&41.2&48.1 \\
		TFA w/ cos~\cite{wang2020frustratingly} & 39.8 & 36.1 & 44.7 & 55.7 & 56.0 & 23.5 & 26.9 & 34.1 & 35.1 & 39.1 & 30.8 & 34.8 & 42.8 & 49.5 & 49.8 \\ 
		MPSR~\cite{wu2020multi} & 41.7 & 42.5 & 51.4 & 55.2 & 61.8 & 24.4 & 29.3 & 39.2 & 39.9 & 47.8 & 35.6 & 41.8 & 42.3 & 48.0 & 49.7 \\ 
		Retentive~\cite{fan2021generalized} & 42.4&45.8&45.9&53.7&56.1&21.7&27.8&35.2&37.0&40.3&30.2&37.6&43.0&49.7&50.1 \\
		FSCE~\cite{sun2021fsce} & 44.2 & 43.8 & 51.4 & 61.9 & 63.4    & 27.3 & 29.5 & 43.5 & 44.2 & 50.2    & 37.2 & 41.9 & 47.5 & 54.6 & 58.5 \\
		Meta FR-CNN~\cite{han2022meta} & 43.0 & {54.5} & {60.6} & {66.1} & {65.4}   & 27.7 & {35.5} & {46.1} & {47.8} & {51.4}   & {40.6} & {46.4} & {53.4} & \underline{59.9} & {58.6} \\ 
		Meta-DETR~\cite{zhang2022metadetr} & 40.6 & {51.4} & {58.0} & 59.2 & {63.6} & \underline{37.0} & {36.6} & {43.7} & {49.1} & \textbf{54.6} & {41.6} & {45.9} & {52.7} & {58.9} & \underline{60.6} \\
		FCT~\cite{han2022fct} & \underline{49.9} & {57.1} & {57.9} & {63.2} & {67.1}    & {27.6} & 34.5 & 43.7 & {49.2} & {51.2}     & {39.5} & \underline{54.7} & {52.3} & {57.0} & {58.7} \\
		VFA~\cite{han2023vfa}&\textbf{57.7}&\textbf{64.6}&\textbf{64.7}&\underline{67.2}&\underline{67.4}&\textbf{41.4}&\textbf{46.2}&\textbf{51.1}&\underline{51.8}&51.6&\textbf{48.9}&\textbf{54.8}&\underline{56.6}&59.0&58.9 \\
		FPD(Ours)&48.1&\underline{62.2}&\underline{64.0}&\textbf{67.6}&\textbf{68.4}&29.8&\underline{43.2}&\underline{47.7}&\textbf{52.0}&\underline{53.9}&\underline{44.9}&53.8&\textbf{58.1}&\textbf{61.6}&\textbf{62.9} \\
		\midrule
		\multicolumn{16}{l}{\textit{Average results over multiple runs:}} \\
		\midrule
		FSDetView~\cite{xiao2020few} &  24.2 & 35.3 &  42.2 &  49.1 &  57.4 & 21.6 & 24.6 &  31.9 &  37.0 &  45.7 & 21.2 &  30.0 &  37.2 &  43.8 &  49.6 \\
		DCNet~\cite{hu2021dense} & 33.9 & 37.4 & 43.7 & 51.1 & 59.6 & 23.2 & 24.8 & 30.6 & 36.7 & 46.6 & 32.3 & 34.9 & 39.7 & 42.6 & 50.7 \\
		Meta-DETR~\cite{zhang2022metadetr}& {35.1} & {49.0} & {53.2} & {57.4} & {62.0} & {27.9} & {32.3} & {38.4} & {43.2} & {51.8} & {34.9} & {41.8} & {47.1} & {54.1} & {58.2} \\
		DeFRCN~\cite{qiao2021defrcn} & {40.2} & \underline{53.6} & {58.2} & {63.6} & \underline{66.5} & \underline{29.5} & \textbf{39.7} & {43.4} & {48.1} & \underline{52.8} & \underline{35.0} & 38.3 & {52.9} & {57.7} & \underline{60.8} \\
		FCT~\cite{han2022fct} & {38.5} & {49.6} & {53.5} & {59.8} & {64.3}  & {25.9} & {34.2} & {40.1} & {44.9} & {47.4}  & {34.7} & {43.9} & {49.3} & {53.1} & {56.3} \\
		VFA~\cite{han2023vfa} & \textbf{47.4}&\textbf{54.4}&\textbf{58.5}&\underline{64.5}&\underline{66.5}&\textbf{33.7}&38.2&\underline{43.5}&\underline{48.3}&52.4&\textbf{43.8}&\textbf{48.9}&\underline{53.3}&\underline{58.1}&60.0 \\
		FPD(Ours)&\underline{41.5}&52.8&\underline{58.4}&\textbf{64.9}&\textbf{67.1}&28.2&\underline{38.7}&\textbf{43.8}&\textbf{50.3}&\textbf{53.6}&34.9&\underline{48.6}&\textbf{54.0}&\textbf{58.4}&\textbf{61.5}  \\
		\bottomrule
	\end{tabular}
	\caption{FSOD results (AP50) on the three splits of Pascal VOC dataset. We report both single run and multiple run results. Bold and Underline indicate the best and the second best results. }
	\label{tab:voc}
\end{table*}

\subsubsection{Test-Time Natural Integration}
A simple method to integrate fine-grained prototypes across different shots is to take the average. However, the detailed features represented by a feature query may not appear in some support images. Directly averaging might hurt the performace. Therefore, we compute a weighted sum using the aforementioned $weight$. Specifically, given $K$ shot support images in a class, which produces $K$ prototypes, the integration is performed via:
\begin{equation}
	p_{avg} = \sum_{s=1}^{K} weight_s^* \cdot p_s
\end{equation}
where $weight^*$ denote the $weight$ after the $softmax$ operation across different shot, $p_{avg}$ is the integrated prototypes. This approach effectively filters out the prototypes that are not compatible with the current feature query, improving the robustness of our detector.

\begin{table}[ht]
	\centering
		\setlength{\tabcolsep}{2.7pt}
		\small
		\begin{tabular}{@{}c|l|l|cc@{}}
			\toprule
			\multirow{2}{*}{} & \multirow{2}{*}{Method} & \multirow{2}{*}{Framework} & \multicolumn{2}{c}{shot} \\
			& & & 10 & 30 \\
			\midrule
			\multicolumn{5}{l}{\textit{Single run results:}} \\
			\midrule
			\multirow{5}{*}{T} &
			TFA w/ cos~\cite{wang2020frustratingly} & FR-CNN & 10.0 & 13.7 \\
			& Retentive~\cite{fan2021generalized} & FR-CNN & 10.5 & 13.8 \\
			&FSCE~\cite{sun2021fsce} & FR-CNN &11.9 & 16.4 \\
			&FADI~\cite{cao2021few}& FR-CNN & 12.2 & 16.1 \\
			& DeFRCN~\cite{qiao2021defrcn} & FR-CNN & \textbf{18.5} & \textbf{22.6} \\
			\midrule
			\multirow{1}{*}{M*} & 
			FCT~\cite{han2022fct} & Transformer & \textbf{17.1}& \textbf{21.4}\\
			\midrule
			\multirow{6}{*}{M} & 
			FSRW~\cite{kang2019few} & YOLOv2 & 5.6 & 9.1 \\
			& Meta R-CNN~\cite{yan2019meta} & FR-CNN & 8.7 & 12.4 \\
			& FSDetView~\cite{xiao2020few} & FR-CNN & 12.5 & 14.7 \\
			& Meta FR-CNN~\cite{han2022meta} & FR-CNN & 12.7 & 16.6 \\
			& VFA~\cite{han2023vfa} & FR-CNN & 16.2 & 18.9 \\
			& FPD(ours) &  FR-CNN & \textbf{16.5} & \textbf{20.1} \\
			\midrule 
			\multicolumn{5}{l}{\textit{Average results over multiple runs:}} \\
			\midrule
			\multirow{2}{*}{T} &
			TFA w/ cos~\cite{wang2020frustratingly} & FR-CNN & 9.1 & 12.1 \\
			& DeFRCN~\cite{qiao2021defrcn} & FR-CNN & \textbf{16.8} & \textbf{21.2} \\
			\midrule
			\multirow{2}{*}{M*} & 
			FCT~\cite{han2022fct} & Transformer & 15.3 & 20.2 \\
			& Meta-DETR~\cite{zhang2022metadetr} & Def DETR & \textbf{19.0} & \textbf{22.2} \\ 
			\midrule
			\multirow{4}{*}{M} & 
			FSDetView~\cite{xiao2020few} & FR-CNN & 10.7 & 15.9 \\
			& DCNet~\cite{hu2021dense} & FR-CNN & 12.8 & 18.6 \\ 
			& VFA~\cite{han2023vfa} & FR-CNN & \textbf{15.9} & 18.4 \\
			& FPD(ours) &  FR-CNN & \textbf{15.9} & \textbf{19.3} \\
			\bottomrule
		\end{tabular}
	\caption{FSOD results (AP) on the MS COCO dataset. T: Transfer-learning based methods. M: Meta-learning based methods. M*: Meta-learning with advanced framework.}
	\label{tab:COCO}
\end{table}

\subsection{High-Level Feature Aggregation}
Feature aggregation between RoI features and class-level prototypes is a crucial step for meta-learning based FSOD, where the high-level semantic information is aligned to make the final prediction. We revisit the conventional methods and propose two improvements from different perspectives. 
\subsubsection{Balanced Class-Agnostic Sampling}
Meta R-CNN adopts a simple class-specific aggregation scheme where the RoI features are aggregated only with the prototypes of the same class. While VFA proposes a class-agnostic aggregation scheme which aggregates RoI features with randomly selected class prototypes to reduce class bias. Nonetheless, we argue that the completely random sampling might disturb the model from focusing on the most crucial positive prototypes and thus hurt the performance. Instead, we propose a balanced sampling strategy named B-CAS which selects a pair of positive and negative prototypes to aggregate with RoI features in parallel. 
The B-CAS not only enables the relation modeling between different classes but also keeps the positive prototype from being overwhelmed by too many negative examples, and therefore can learn the high-level semantic relations more effectively.

~\cite{fan2020few} employs a more complex training strategy which divides training pairs into three types and maintains a ratio of 1:2:1. Additionlly, a matching loss is computed to align RoI features with prototypes. 
However, we find it instead hurts the performance. A plausible reason is that FFA introduces the asymmetry upon two branches, making the matching loss no longer beneficial. 
Consequently, a simple yet effective method B-CAS is adopted in our experiments.

\subsubsection{Non-Linear Fusion Module}
Many previous meta-learning based methods use element-wise multiplication to handle the feature fusion. 
We argue that while this approach learns the similarities within the same class effectively, it struggles to capture the class differences. Therefore it is not compatible with the proposed B-CAS. To solve this problem, we employ a novel non-linear fusion network following~\cite{han2022meta, xiao2020few} with modifications.

Specifically, features after element-wise multiplication, subtraction and concatenation are processed independently to refine their relation to the new feature. Then they are concatenated with the vanilla RoI features and further refined before fed into the detection head. Given RoI feature $f_{roi} \in \mathbb{R}^{1 \times 2d}$ and class prototype $p_{cls} \in \mathbb{R}^{1 \times 2d}$, the aggregation can be formulated as:
\begin{equation}
	f^{'}= [\mathcal{F} _{1}(f_{roi} \odot p_{cls}), \mathcal{F}_{2}(f_{roi} - p_{cls}), \mathcal{F}_{3}[f_{roi}, p_{cls}], f_{roi}]
\end{equation}
\begin{equation}
	f = \mathcal{F}_{agg}(f^{'})
\end{equation}
where $\mathcal{F}_{1}$, $\mathcal{F}_{2}$ and $\mathcal{F}_{3}$ represent independent fully-connected layer followed by ReLU activation function, and $\mathcal{F}_{agg}$ denote a pure fully-connected layer. This formulation provides a stronger capability to thoroughly explore the relations between high-level features. In addition, an exclusive path for RoI features is reserved to propagate the original RoI information, which reduces the noise introduced by random prototypes and can be used to regress the object location.

\section{Experiments}
\subsection{Benchmarks}
We evaluate our method on two widely-used FSOD benchmarks PASCAL VOC~\cite{everingham2010pascal} and MS COCO~\cite{lin2014microsoft}, using exactly the same class partitions and few-shot examples as in~\cite{wang2020frustratingly}.

\noindent
\textbf{PASCAL VOC. }
The 20 PASCAL VOC classes are split into 15 base classes and 5 novel classes. There are three different class partitions for a more comprehensive evaluation. The VOC07 and VOC12 train/val sets are used for training and the VOC07 test set is used for evaluation. The Mean Average Precision at IoU=0.5 (AP50) is reported under $K$=\{1, 2, 3, 5, 10\} shot settings. 

\noindent
\textbf{MS COCO. } For MS COCO, the 20 PASCAL VOC classes are used as novel classes, the other 60 classes are used as base classes. The 5k images from COCO2017 val are used for evaluation and the rest are used for training. We report the AP at IoU=0.5:0.95 under $K$=\{10, 30\} shot settings. 

\begin{table}[t]
	\centering
		\begin{tabular}{lccc|ccc}
			\toprule
			\multirow{2}{*}{} & \multirow{2}{*}{B-CAS} & \multirow{2}{*}{NLF} & \multirow{2}{*}{FFA} & \multicolumn{3}{c}{shot} \\
			& & & & 3 & 5 & 10 \\
			\midrule
			Baseline&&&&56.7&58.3&61.4 \\
			\midrule
			\multirow{3}{*}{Ours}&$\checkmark$&&&61.2&64.7&64.9 \\
			&$\checkmark$&$\checkmark$&&62.8&67.1&66.3 \\
			&$\checkmark$&$\checkmark$&$\checkmark$&\textbf{64.0}&\textbf{67.6}&\textbf{68.4} \\
			\bottomrule
		\end{tabular}
	\caption{Ablation study of different components. }
	\label{tab:ablation_1}
\end{table}

\subsection{Implementation Details}
Our method is implemented with MMDetection~\cite{mmdetection}. We adopt ResNet-101~\cite{he2016resnet} pretrained on ImageNet~\cite{russakovsky2015imagenet} as the backbone. The single scale feature map is used for detection without FPN~\cite{lin2017feature}. 
We resize the query images to a maximum of 1333x800 pixels, and the cropped instances from support images are resized to 224x224 pixels.

Our model is trained on 2x3090 Nvidia GPUs with a total batch size of 8, using the SGD optimizer. In the base training stage, the model is trained on VOC and COCO datasets for 20k/110k iterations. The learning rate is set to 0.004 and decayed at 17k/92k iteration by a factor of 0.1.
In the fine-tuning stage, the learning rate is set to 0.001. We use exactly the same loss functions with Meta R-CNN.

\subsection{Comparison with the State-of-the-Art Methods}
\noindent
\textbf{PASCAL VOC.} We show both the single run results and the average results over multiple runs of PASCAL VOC in Table~\ref{tab:voc}. It can be seen that FPD significantly outperforms previous methods, achieving the state-of-the-art performance in most settings. Specifically, FPD outperforms previous best method by 1.5\%, 4.4\%, and 6.8\% on the three data splits under $K$=10 shot setting, respectively. We notice that under $K$=\{1, 2\} shot settings, our method is less effective than VFA, which is a strong FSOD detector utilizing a variational autoencoder to estimate class distributions. 
Our analysis suggests that in extremely data-scarce scenarios, it is more challenging for the FFA to capture the representative and common features across different shots, therefore it fails to achieve the expected effect under $K$=\{1, 2\} shot settings. 

\noindent
\textbf{MS COCO.} Table~\ref{tab:COCO} shows the results of MS COCO.
It can be seen that FPD outperforms all of the meta-learning based methods adopting the Faster R-CNN framework. For example, FPD improves performance by 6.3\% compared to previous best result under $K$=30 shot setting. FPD ranks fourth among all the methods. Please note that our method focuses on the three proposed components, without using advanced frameworks or techniques such as DETR, Transformer or gradient decoupled layer. Given the challenging nature of the MS COCO dataset, we believe that the performance can be further improved with more refinements. 

\begin{figure*}[ht]
	\centering
	\includegraphics[width=0.88\linewidth]{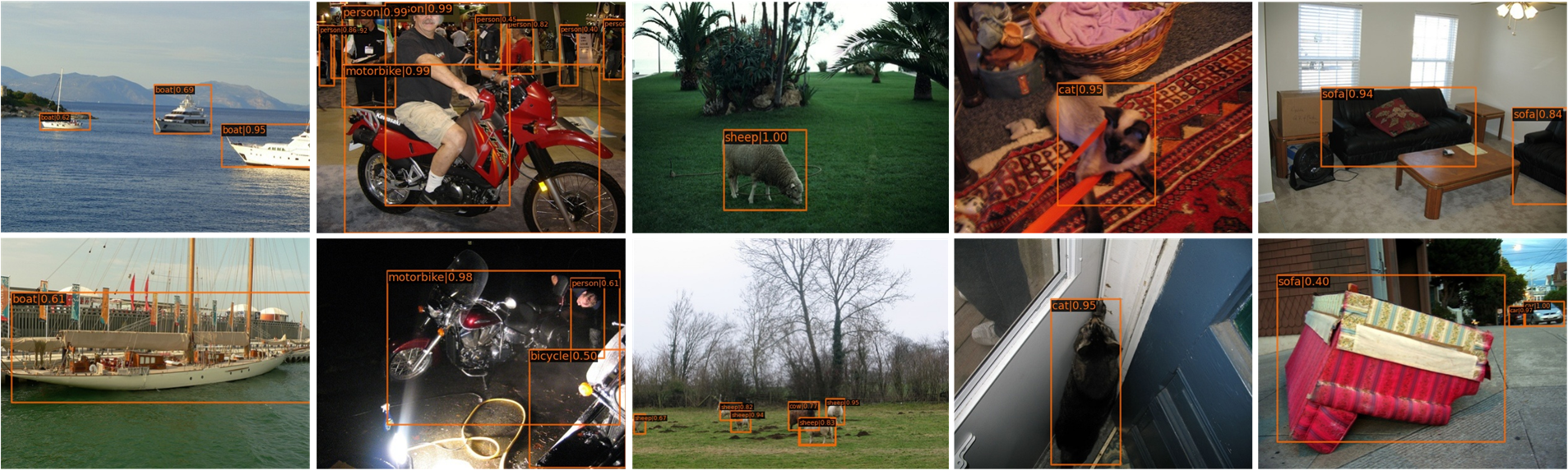}
	\caption{Visualization of the detection results on novel classes.  }
	\label{fig:detection_results}
\end{figure*}

\begin{table}[ht]
	\centering
	\adjustbox{width=\linewidth}{
		\begin{tabular}{lcc|ccc}
			\toprule
			\multirow{2}{*}{Method} & \multirow{2}{*}{Directly Match} & \multirow{2}{*}{FFA} & \multicolumn{3}{c}{shot} \\
			& & & 3 & 5 & 10 \\
			\midrule
			Baseline*&&&62.8&67.1&66.3 \\
			\midrule
			\multirow{2}{*}{Ours}&$\checkmark$&&63.2&67.0&67.5 \\
			&&$\checkmark$&\textbf{64.0}&\textbf{67.6}&\textbf{68.4} \\
			\bottomrule
	\end{tabular}}
	\caption{Comparison with directly matching. }
	\label{tab:ablation_2}
\end{table}

\subsection{Ablation Study}
We conduct comprehensive experiments on the Novel Set 1 of PASCAL VOC under $K$=\{3, 5, 10\} shot settings, which demonstrates the effectiveness of our proposed method.

\noindent
\textbf{Effect of Different Components.} 
We show the results with different components in Table~\ref{tab:ablation_1}. It can be seen that B-CAS and NLF together improve the performance by about 10\% over the baseline. Based on this, our FFA can further boost the results, achieving the state-of-the-art performance.

\noindent
\textbf{Effect of the FFA.} 
FFA differs from DCNet in that it distills the fine-grained prototypes to aggregate with query branch. To demonstrate the superiority of this method, we re-implement the DRD module following DCNet to directly match dense feature maps for aggregation. We show the experimental results in Table~\ref{tab:ablation_2}. It can be seen that FFA consistently achieves better performance than directly matching, which validates the effectiveness of our method.

\begin{figure}[h]
	\centering
	\includegraphics[width=0.87\linewidth]{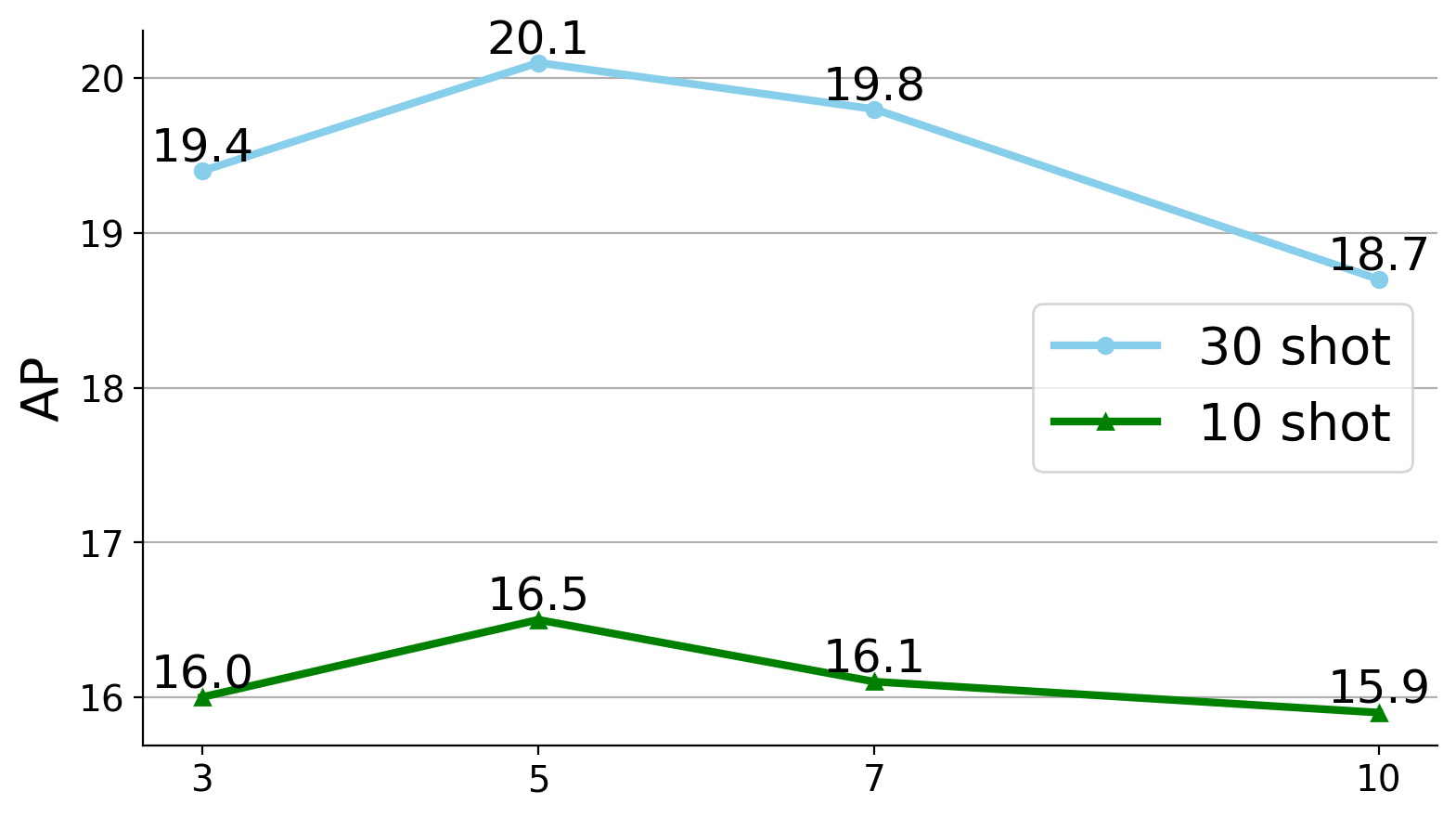}
	\caption{Ablation study on the number of feature quries. }
	\label{fig:num_queries}
\end{figure}

\noindent
\textbf{Effect of Feature Queries. } 
We assign each class a set of feature queries, which are the key guidance to distill fine-grained prototypes. The number of feature queires for a class is set to 5 by default. Figure~\ref{fig:num_queries} shows the effect of this number.

\begin{figure}[h]
	\centering
	\includegraphics[width=0.95\linewidth]{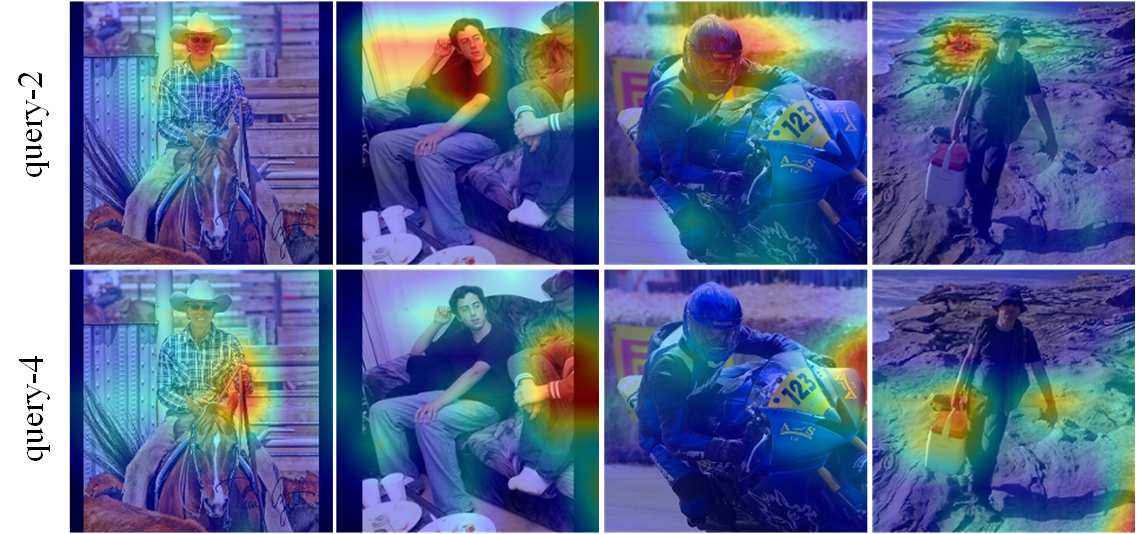}
	\caption{Attention heatmap of feature queries. Please find more discussion and results in Appendix.}
	\label{fig:query_00}
\end{figure}

Moreover, to explore the fundamental working machanism, we visualize the attention heatmap of feature queries on support images. As shown in Figure~\ref{fig:query_00}, two feature queries from \textit{person} category are listed. They are prone to focus on the specific details, e.g., \textit{head} and \textit{hand}, which conforms to our expectations. Please note that the generated heat maps has a resolution of 14x14. It is not absolutely aligned with the original images. 

\subsection{Visualize Detection Results}
We show the detection results in Figure~\ref{fig:detection_results}. The model is trained on the Novel Set 3 of PASCAL VOC under 10 shot setting and tested on the VOC07 test set. It can be seen that many of the novel instances are effectively detected, even though the detected bboxes are not perfectly aligned. This results demonstrate the promising potential of our method.

\section{Conclusion}
This paper studies the meta-learning based FSOD. We propose a novel FFA module which can distill fine-grained prototypes in addition to class-level ones. It enables more robust novel object detection by focusing on the detailed features. We also propose B-CAS strategy and NLF module to aggregate high-level features more effectively. Both quantitative and qualitative results demonstrate the effectiveness of our method and the promising prospect of FSOD.

\section{Acknowledgments}
This work was supported in part by the Overseas Students Science and Technology Activities Project (No. 2018024), 
by the National Natural Science Foundation of China (No. 61502389), by the Natural Science Basic Research Program of Shaanxi Province, China (No. 2023-JC-YB-508). 

\bibliography{aaai24}

\clearpage
\appendix
\section{Appendix}

\subsection{Additional Visualization}
\begin{figure}[ht]
	\centering
	\includegraphics[width=\linewidth]{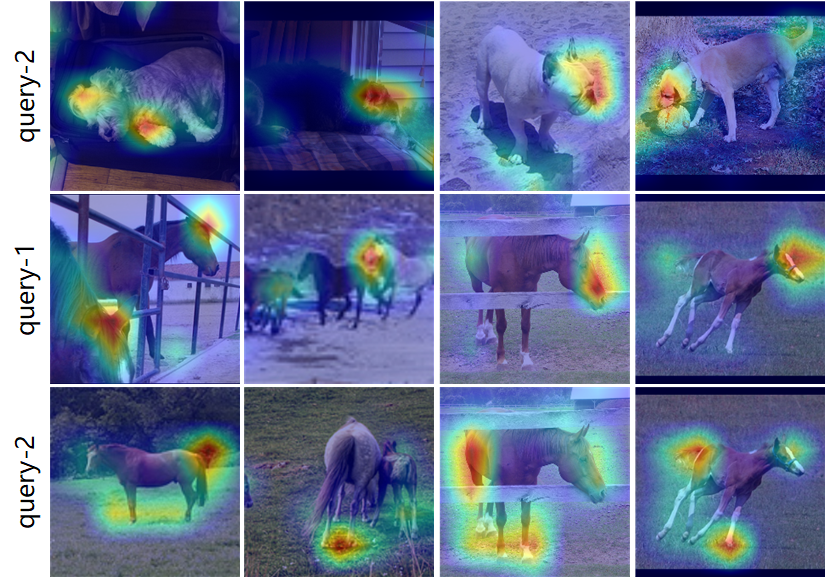}
	\caption{Additional attention heatmap of feature queries. The model is trained on Novel Set 3 of PASCAL VOC.}
	\label{fig:query_03}
\end{figure}

\textbf{Attention Heatmap of Feature Queries. } We show more attention heatmaps of feature queries upon support images in Figure~\ref{fig:query_03}. 
We can see that the feature query 2 from \textit{dog} category is prone to capture the detailed features of \textit{head}. The feature query 1, 2 from \textit{horse} category are focus on \textit{head} and \textit{legs}, respectively. 
The feature queries are more likely to capture the different details, rather than collapse to a trivial solution. 

\begin{figure}[ht]
	\centering
	\includegraphics[width=\linewidth]{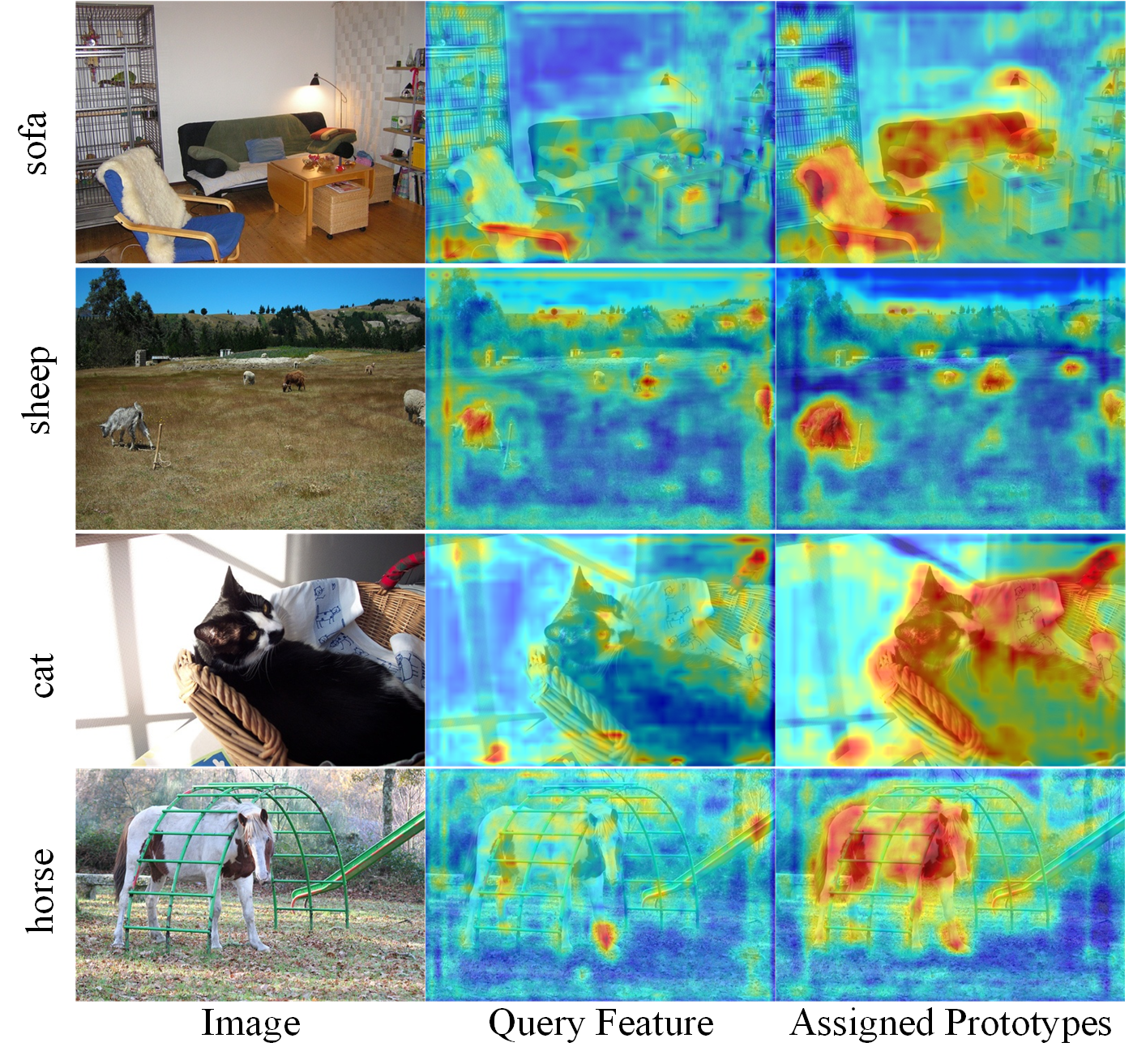}
	\caption{Feature map of query images. }
	\label{fig:query_feature_map}
\end{figure}

\noindent
\textbf{Feature Map of Query Images. } The feature map of a query image $X_q \in \mathbb{R}^{HW \times d}$ are summed alone dimension $d$ and then normalized to $[0, 1]$ to produce the heatmap. We show the results of original query features and the assigned prototypes in Figure~\ref{fig:query_feature_map}. It can be seen that the assigned prototypes can highlight the representative features to facilitate the model prediction. 
All these evidences demonstrate the effectiveness of our proposed FFA. 

\subsection{Additional Implementation Details}
Our method follows the two-stage training paradigm. At the base training stage, we train all of the model parameters (the first few layers of ResNet are freezed conventionally). At the fine-tuning stage, we freeze the backbone and only train the RPN, FFA and NLF module. Fine-tuning of the FFA together with RPN can help to produce high-quality proposals of the novel classes. Under $K$=\{1, 2\} shot settings, we freeze the RPN to avoid overfitting.

\subsection{Computational Cost}
Table~\ref{tab:inference_time_2} shows the computational cost of different methods at inference time. We conduct the experiments on a single Nvidia 3090 GPU. The batch size is set to 1. It can be seen that our method has a better trade-off between the performance and computational efficiency.
\begin{table}[ht]
	\centering
	\small
	\setlength{\tabcolsep}{2.2pt}
	\begin{tabular}{@{}l|l|c|c|c@{}}
		\toprule
		Dataset & Method & Params(MB) & FLOPs(GB) & FPS(img/s)  \\
		\midrule
		\multirow{3}{*}{\parbox{1.2cm}{VOC\\(20 class)}} & Baseline & 45.99 &  709.76 & 16.2 \\
		& FPD(Ours) & 65.68 & 818.10 & 14.8 \\
		& Directly Match & 69.58 & 956.72 & 14.5 \\
		\midrule
		\multirow{3}{*}{\parbox{1.2cm}{COCO\\(80 class)}} & Baseline & 46.72 & 766.36 & 7.3 \\
		& FPD(Ours) & 66.5 & 1309.50 & 6.5 \\
		& Directly Match & 70.32 & 1466.25 & 5.3 \\
		\bottomrule
	\end{tabular}
	\caption{The computational cost at the inference time.}
	\label{tab:inference_time_2}
\end{table}

\begin{figure}[ht]
	\centering
	\includegraphics[width=\linewidth]{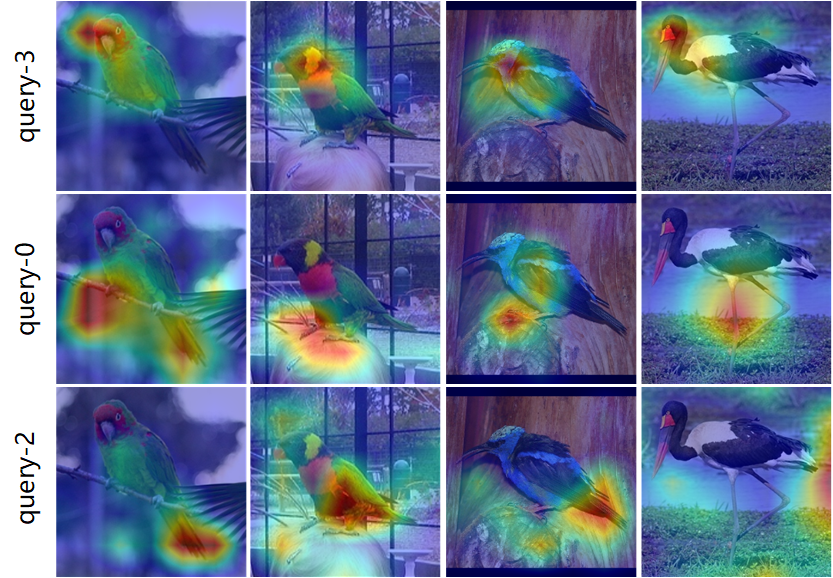}
	\caption{Attention heatmap of feature queries (bird).}
	\label{fig:query_attention_bird}
\end{figure}

\begin{figure}[h]
	\centering
	\includegraphics[width=\linewidth]{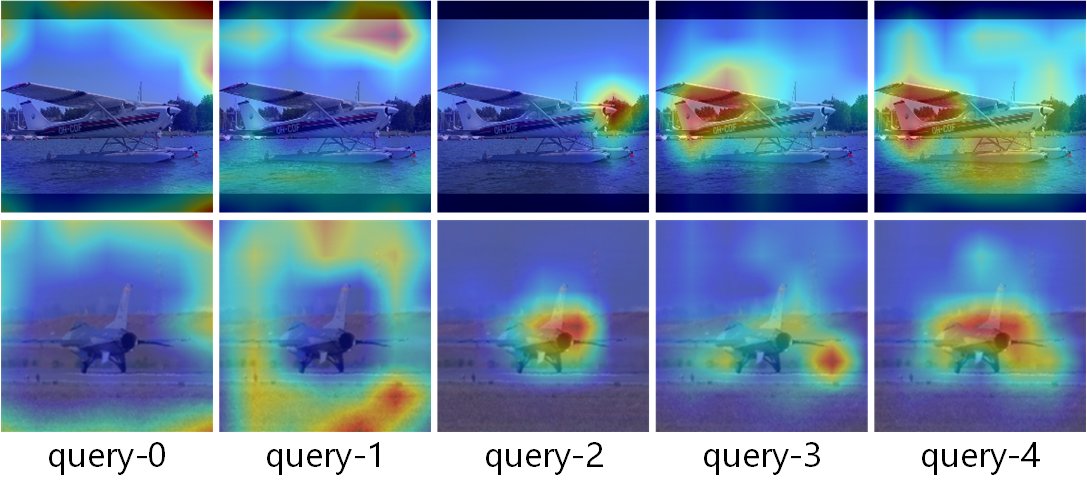}
	\caption{Attention heatmap of feature queries (airplane). }
	\label{fig:query_attention_airplane}
\end{figure}

\section{More Discussion}
Our proposed FFA module has similarities with DCNet and Meta-DETR. In this part, we provide a more detailed comparison among these methods.
\subsection{Compare with DCNet}
\begin{figure}[h]
	\centering
	\includegraphics[width=0.96\linewidth]{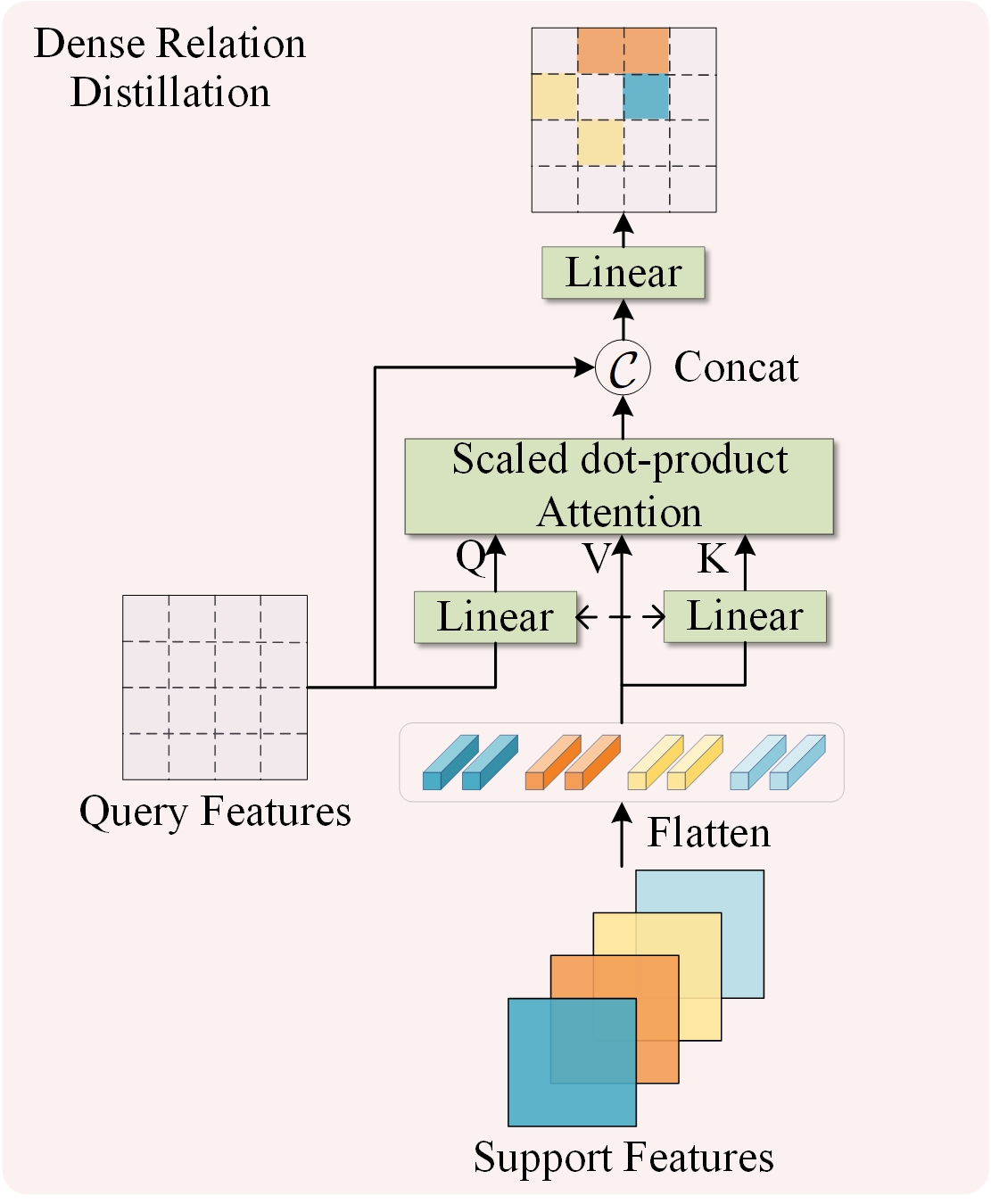}
	\caption{The Dense Relation Distillation module of DCNet. }
	\label{fig:drd_}
\end{figure}

Figure~\ref{fig:drd_} illustrates the DRD module of DCNet, which densely matches all classes of support features into the query feature map. There are two main differences between DRD and our FFA (as shown in Figure~\ref{fig:FFA}).
\textbf{First}, FFA utilizes feature queries to distill fine-grained prototypes, enabling the model to focus on the most representative detailed features and to reduce computational costs (see Table~\ref{tab:inference_time_2}). It also enhances inference efficiency (see subsec. Test-Time Natural Integration). \textbf{Second}, FFA employs a residual connection for the original query features, and the prototypes are directly assigned to the query feature map without any extra projection. This maintains the query-support branches in the same feature space, which is crucial for the subsequent high-level feature fusion operation.

\raggedbottom
\subsection{Compare with Meta-DETR}
\begin{figure}[h]
	\centering
	\includegraphics[width=0.94\linewidth]{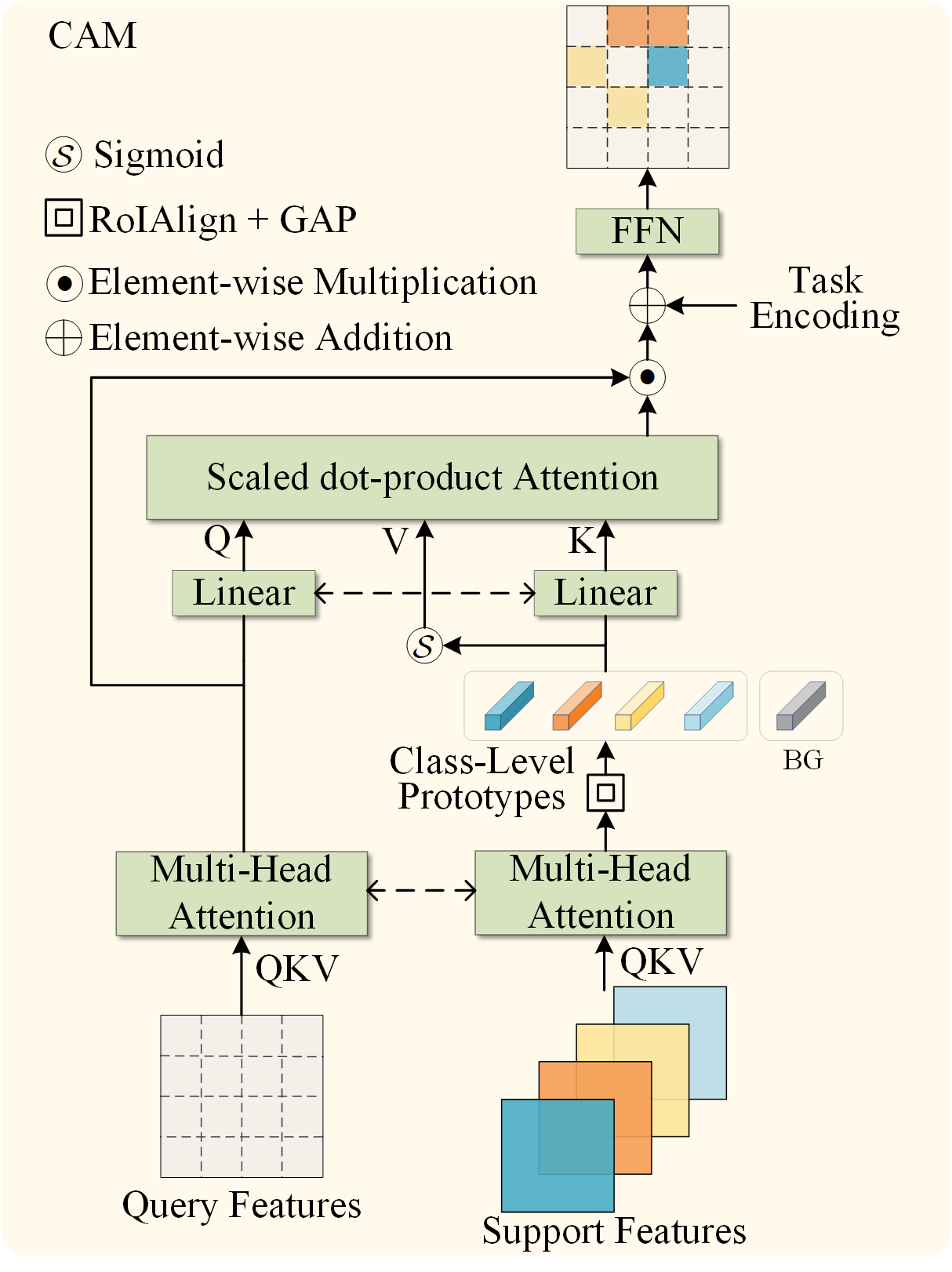}
	\caption{The Correlational Aggregation Module of Meta-DETR.}
	\label{fig:cam_}
\end{figure}

Meta-DETR incorporates meta-learning and attention mechanism into the DETR framework.
It utilizes the cross attention operation to aggregate query-support features. 
As shown in Figure~\ref{fig:cam_}, CAM performs global average pooling to generate the class-level prototypes. They are matched with query features and then assigned into query features based on the matching results. Instead of performing element-wise addition, the element-wise multiplication operation is used to rewight the query feature map along the channel dimension. 

CAM differs from our method in three main aspects. \textbf{First}, it focuses on high-level feature aggregation, while our FFA is used to aggregate detailed features. FFA utilizes feature queries and an additional cross attention layer to refine the important local context into the fine-grained prototypes.  \textbf{Second}, CAM employs sigmoid and multiplication operations to reweight the query feature map, while FFA directly adds the assigned prototypes to it, preserving more information and potential in the early stages. \textbf{Third}, CAM incorporates a novel and effective encoding matching task to predict object classes.

\subsection{Revised Performance}
After carefully re-examing our code, we found some unintentional discrepancies that have impacted the performance metrics. These mistakes do not compromise the main contributions of this work. Table~\ref{tab:voc_revised} shows the revised results.

\begin{table*}[ht]
	\centering
	\setlength{\tabcolsep}{4.6pt}
	\small
	\begin{tabular}{@{}l|ccccc|ccccc|ccccc@{}}
		\toprule
		\multirow{2}{*}{Method / shot} & \multicolumn{5}{c|}{Novel Set 1} & \multicolumn{5}{c|}{Novel Set 2} & \multicolumn{5}{c}{Novel Set 3} \\
		& 1 & 2 & 3 & 5 & 10 & 1 & 2 & 3 & 5 & 10 & 1 & 2 & 3 & 5 & 10 \\
		\midrule
		\multicolumn{16}{l}{\textit{Single run results:}} \\
		\midrule
		FSCE~\cite{sun2021fsce} & 44.2 & 43.8 & 51.4 & 61.9 & 63.4    & 27.3 & 29.5 & 43.5 & 44.2 & 50.2    & 37.2 & 41.9 & 47.5 & 54.6 & 58.5 \\
		Meta FR-CNN~\cite{han2022meta} & 43.0 & {54.5} & {60.6} & {66.1} & {65.4}   & 27.7 & {35.5} & {46.1} & {47.8} & {51.4}   & {40.6} & {46.4} & {53.4} & \underline{59.9} & {58.6} \\ 
		Meta-DETR~\cite{zhang2022metadetr} & 40.6 & {51.4} & {58.0} & 59.2 & {63.6} & \underline{37.0} & {36.6} & {43.7} & {49.1} & \underline{54.6} & {41.6} & {45.9} & {52.7} & {58.9} & \underline{60.6} \\
		FCT~\cite{han2022fct} & \underline{49.9} & {57.1} & {57.9} & {63.2} & {67.1}    & {27.6} & 34.5 & 43.7 & {49.2} & {51.2}     & {39.5} & \underline{54.7} & {52.3} & {57.0} & {58.7} \\
		VFA~\cite{han2023vfa}&\textbf{57.7}&\textbf{64.6}&\underline{64.7}&\underline{67.2}&\underline{67.4}&\textbf{41.4}&\textbf{46.2}&\textbf{51.1}&\underline{51.8}&51.6&\textbf{48.9}&\textbf{54.8}&\underline{56.6}&59.0&58.9 \\
		\textbf{FPD(Previous)}&{46.5}&\underline{62.3}&\textbf{65.4}&\textbf{68.2}&\textbf{69.3}&32.2&\underline{43.6}&\underline{50.3}&\textbf{52.5}&\textbf{56.1}&\underline{43.2}&53.3&\textbf{56.7}&\textbf{62.1}&\textbf{64.1}   \\
		\textbf{FPD(Revised)}&48.1&62.2&64.0&67.6&68.4&29.8&43.2&47.7&52.0&53.9&44.9&53.8&58.1&61.6&62.9 \\
		
		\bottomrule
	\end{tabular}
	\caption{Revised FSOD results (AP50) on the three splits of Pascal VOC dataset. }
	\label{tab:voc_revised}
\end{table*}

\end{document}